\documentclass[letterpaper,10pt,conference]{ieeeconf}

\IEEEoverridecommandlockouts
\usepackage{cite}
\usepackage{amsmath,amssymb}
\usepackage{algorithmic}
\usepackage{graphicx}
\usepackage{textcomp}
\usepackage{xcolor}
\usepackage{booktabs}
\usepackage{array}
\usepackage{balance}
\usepackage{tabularx}
\usepackage{xspace}

\usepackage{multirow}
\usepackage{makecell}

\usepackage{cuted}    
\usepackage{capt-of}  
\usepackage{url}

\usepackage{makecell}
\renewcommand\theadfont{\normalfont} 

\newcommand{\method}{\textit{FWIK}\xspace}

\usepackage{authblk}

\title{\LARGE\bfseries{\textit{\method}: Multi-Waypoint IK Selection via Lookahead Graph Search}}

\begin{document}

\author{Jinglan Cao, Guohui Zhong, and Chengwei Zhang}

\maketitle

\thispagestyle{empty}
\pagestyle{empty}
\begin{strip}
  \vspace{-35pt}
  \centering
  \includegraphics[width=\textwidth,keepaspectratio]{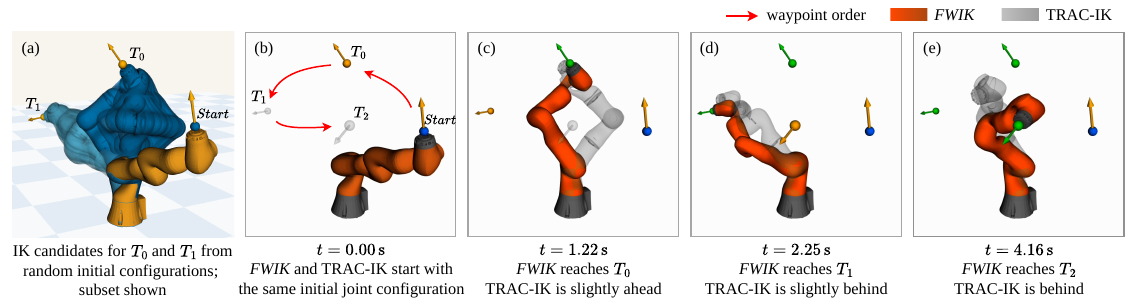}
  \captionof{figure}{
        We present a method that enables a robot to select IK solutions for an ordered sequence of waypoints by reasoning over multiple candidate configurations in a bounded future window.
        (a) Candidate joint configurations generated for upcoming waypoints. (b) Experimental setup starting from identical initial states; red arrows show waypoint order. (c)--(e) Time-series comparison demonstrating that our method reaches subsequent waypoints faster than TRAC-IK. Orange and gray represent \method and TRAC-IK, respectively.}
  \label{fig:fwik_tracik_motion_sequence}
  \vspace{-5pt}
\end{strip}

\begin{abstract}
An ordered sequence of end-effector poses can give a redundant robot arm multiple valid joint configurations at each waypoint. Inverse kinematics (IK) computes these configurations. Choosing a configuration only for the next waypoint can leave the arm poorly placed for later waypoints. This may increase planned motion time or cause a later planning failure. This paper presents Future-Window Inverse Kinematics (\method), an online method that selects IK solutions by considering a small window of future waypoints. \method generates several IK candidates, connects them in a layered joint-space graph, and ranks candidate paths with a simple motion-time estimate. A learned model removes paths whose transitions are likely to cause self-collision. A standard MoveIt~2 planner then checks the next-waypoint IK goals from the remaining paths in ranked order, and the first goal that yields a valid trajectory is selected for execution. We evaluate \method in simulation on two 7-DoF manipulators, the Franka Emika Panda and the KUKA LBR iiwa. Each method-planner setting contains five repetitions of 100 sequences with 50 waypoints. Across Pilz, RRT-Connect, CHOMP, and STOMP, \method completes 96.8--100\% of sequences and gives the shortest planned trajectory duration in all eight robot-planner settings. Its mean duration is 12.8--13.1\% lower than STAMPEDE, 22.5--25.6\% lower than IKLink, and 27.5--37.2\% lower than retry-based TRAC-IK. The results show that a small future window can improve sequence completion and reduce planned motion time while leaving final trajectory validation to the motion planner.
\end{abstract}


\section{Introduction}
Many robot tasks require an ordered sequence of end-effector poses rather than a single isolated goal. Welding, surface inspection, and multi-stage pick-and-place are common examples. For a redundant manipulator, the same end-effector pose may admit multiple valid joint configurations because the robot has more degrees of freedom (DoF) than are required to specify the end-effector pose~\cite{siciliano2009robotics}. Inverse kinematics (IK) determines such configurations for a desired end-effector pose.

The choice of IK solution affects more than the current motion. The selected configuration becomes the start state for the next planning request, so locally equivalent IK solutions can have very different consequences for the remaining waypoint sequence. One configuration may preserve joint range and maintain connectivity to future IK branches, whereas another may require a large posture change or make a later waypoint difficult to reach. Repeated local choices can therefore increase planned trajectory duration, introduce discontinuous motion, or cause the sequence to terminate when a later waypoint cannot be planned~\cite{ames2022ikflow,wang2023rangedik,yoon2023learning}.

Standard online IK interfaces are typically pointwise: given a target pose, they return a feasible configuration for that pose without explicitly reasoning about downstream waypoints. A solver such as TRAC-IK can quickly provide a valid configuration under joint limits~\cite{beeson2015tracik}, after which a motion planner generates a trajectory toward that goal. This separation is simple and efficient, but it discards useful future information when several upcoming waypoints are already known. 
Methods that generate diverse IK solutions or optimize configurations over an entire path address related aspects of this problem~\cite{ames2022ikflow,rakita2019stampede,wang2024iklink,kang2022torm,morgan2024cppflow}. However, full-path methods exploit the complete known sequence before execution, whereas an online system may instead benefit from bounded lookahead: using only a short future window, executing one decision, and updating that decision as the task progresses or new information becomes available.

Self-collision introduces a second source of inefficiency. Geometric collision checkers such as the Flexible Collision Library (FCL) provide accurate collision tests and remain essential for final validation~\cite{pan2012fcl}. However, if the IK selection layer proposes a goal whose transition from the current configuration is likely to self-collide, the downstream planner may spend substantial effort on an unfavorable target before failing and retrying another IK solution. Learning-based collision models have shown that approximate configuration-space representations can reduce the cost of repeated collision queries~\cite{rakita2018relaxedik,das2020learning,zhi2022diffco}. For multi-waypoint execution, however, configuration validity alone is insufficient: the relevant question is also whether the transition from the current state to a candidate IK solution is likely to be collision-prone.

These observations motivate the central question of this paper: \textit{Can finite-horizon IK selection with edge-level self-collision screening improve the success--time trade-off of online multi-waypoint manipulation compared with pathwise or retry-based IK selection?}

We answer this question with Future-Window Inverse Kinematics (\method), an online IK decision layer that operates upstream of a standard motion planner. At each execution step, \method constructs a layered graph over a finite window of upcoming waypoints. Each node represents an IK candidate, and each edge represents a possible transition between two candidates with an associated synchronized joint-motion time estimate. Rather than selecting the next IK target independently, \method ranks candidate paths according to their connectivity and estimated motion cost across the visible window. The leading transitions of promising paths are then screened using a learned edge-level self-collision model before distinct first-goal candidates are submitted to a MoveIt 2 planner for geometric validation. Only the first validated IK target is executed; the window then advances by one waypoint and the decision process is repeated. In this way, \method uses future information without replacing the downstream motion planner or committing to a complete trajectory in advance.

The main contributions of this paper are:
\begin{itemize}
\item A finite-window formulation of online IK selection that chooses the next IK target according to its connectivity with near-future waypoint solutions, rather than considering the next pose in isolation.
\item An edge-level self-collision screening module that rejects risky leading transitions before costly planner retries while retaining geometric collision checking in the downstream planning pipeline.
\item A repeated evaluation on two manipulators and four motion planners, comparing \method against IKLink, STAMPEDE, and TRAC-IK in terms of sequence success, waypoint completion, IK decision time, and planned trajectory duration.
\end{itemize}

\section{Related Work}

\method lies between pointwise IK and full-path IK. It uses a bounded amount of future waypoint information to select the next IK target, while retaining a conventional motion planner for downstream trajectory generation and validation. This section reviews the related work in IK generation and pathwise IK, motion planning, and collision-aware transition ranking. Each subsection then positions \method within these areas.

\subsection{Pointwise and Pathwise Inverse Kinematics}

Classical inverse kinematics maps a Cartesian end-effector pose to one or more joint configurations while accounting for kinematic constraints such as singularities and joint limits~\cite{nakamura1986singularity}. TRAC-IK improves the robustness of pointwise IK by combining inverse-Jacobian search with numerical optimization~\cite{beeson2015tracik}. RelaxedIK formulates real-time IK as a multi-objective optimization that balances pose tracking with motion-related objectives, including smoothness, self-collision avoidance, and singularity avoidance~\cite{rakita2018relaxedik}. IKFlow instead learns to generate diverse joint configurations for a single target pose~\cite{ames2022ikflow}, while RangedIK extends IK optimization to targets defined over ranges of valid poses~\cite{wang2023rangedik}. These approaches improve the generation or optimization of IK solutions for individual targets, but they do not directly address online branch selection when multiple future waypoints are already known.

Pathwise IK methods reason over multiple target poses jointly. STAMPEDE constructs a discrete graph of candidate configurations across a pose sequence~\cite{rakita2019stampede}, while IKLink combines graph search and dynamic programming to reduce joint-space reconfiguration across successive targets~\cite{wang2024iklink}. TORM and CppFlow similarly optimize or generate joint-space realizations for complete Cartesian paths~\cite{kang2022torm,morgan2024cppflow}. More recent work considers many-objective formulations and model-predictive approaches for joint-space path generation~\cite{xie2025many,oelerich2025boundmpc}.

\method shares the graph-based view of pathwise IK, but differs in its execution contract. Rather than optimizing a complete waypoint sequence before execution, it constructs a finite future window, ranks candidate paths within that window, validates candidate goals for the current step, executes a single IK target, and then recomputes as the window advances. This receding-horizon formulation preserves useful short-term branch connectivity while allowing the IK decision to be updated after every executed waypoint.

This problem is also distinct from robotic task sequencing. Task-sequencing methods typically optimize the order in which task locations are visited~\cite{alatartsev2015task,suarez2018robotsp}. In contrast, \method assumes that the waypoint order is fixed; its decision variable is the joint-space IK branch used to realize each waypoint.

\subsection{Motion Planning and Planner Validation}

Motion planning determines a collision-free trajectory between robot configurations in high-dimensional configuration spaces. Sampling-based approaches such as probabilistic roadmaps and rapidly exploring random trees form a major class of such methods~\cite{kavraki1996prm,kuffner2000rrt}. Among these, RRT-Connect is widely used for dual-tree bidirectional search in high-dimensional query tasks~\cite{kuffner2000rrt}. The Open Motion Planning Library (OMPL) provides reusable implementations of many sampling-based planners~\cite{sucan2012ompl}, while MoveIt integrates inverse kinematics, motion planning, collision checking, and trajectory execution within a common manipulation framework~\cite{coleman2014moveit}. For structured industrial motions, the Pilz Industrial Motion Planner provides synchronized point-to-point joint trajectories together with linear and circular Cartesian motions~\cite{pilz2018industrial}.

Optimization-based planners provide a complementary approach. CHOMP optimizes a trajectory objective that balances smoothness and obstacle avoidance~\cite{zucker2013chomp}; STOMP improves trajectories through stochastic sampling and iterative updates~\cite{kalakrishnan2011stomp}; and TrajOpt formulates trajectory optimization using sequential convex optimization with collision penalties~\cite{schulman2014trajopt}.

\method does not replace these planners. Instead, it operates upstream of motion planning by ranking alternative IK goals for the next waypoint using information from nearby future waypoints. The configured planner remains responsible for trajectory construction, geometric collision checking, and final validation. The proposed method therefore addresses goal-configuration selection rather than trajectory generation itself, allowing the selector to remain compatible with different downstream planning pipelines.

\subsection{Collision Checking and Time-Aware Ranking}

Collision checking is a major computational component of robot motion planning~\cite{zhi2022diffco,koptev2023neuraljsdf,li2024rdf}. Geometric libraries such as FCL provide accurate collision and proximity queries and remain important for validating planned robot motion~\cite{pan2012fcl}. Learning-based approximations seek to reduce repeated geometric collision queries by learning configuration-space or distance representations. Fastron, for example, learns an approximate collision boundary for fast configuration classification~\cite{das2020learning}, while later approaches learn differentiable or neural representations of collision and signed-distance structure~\cite{zhi2022diffco,koptev2023neuraljsdf,li2024rdf}.

For waypoint execution, however, endpoint validity alone is insufficient. A candidate IK configuration may itself be collision-free while the transition from the current configuration to that candidate passes through self-collision. This motivates evaluating motion segments rather than only isolated configurations, consistent with collision-aware trajectory optimization and learned collision reasoning over local motion structure~\cite{schulman2014trajopt,rakita2021collisionik,zhi2022diffco}. \method adopts this edge-level view and uses a learned transition classifier to screen risky leading motions before they are submitted to the downstream planner. The learned model acts only as a scoring and rejection mechanism; geometric planner validation is retained for the selected goal.

Transition ranking also requires a lightweight estimate of motion cost. Time-aware trajectory generation commonly accounts for joint velocity and acceleration limits when estimating or constructing feasible motions~\cite{kroger2009online,kunz2012time}. Building on this principle, \method uses a synchronized trapezoidal joint-motion model to estimate the traversal time of graph edges. This provides a computationally inexpensive score for comparing alternative IK branches before invoking the full motion planner.

Together, these lines of work motivate a middle ground between pointwise IK selection and complete path optimization: a receding-horizon IK decision layer that exploits near-future branch connectivity, screens transition-level self-collision risk, and leaves exact trajectory generation and validation to an existing motion planner.

\section{Methodology}

\begin{figure}[t]
    \centering
    \includegraphics[
    width=\columnwidth,
    height=0.42\textheight,
    keepaspectratio
    ]{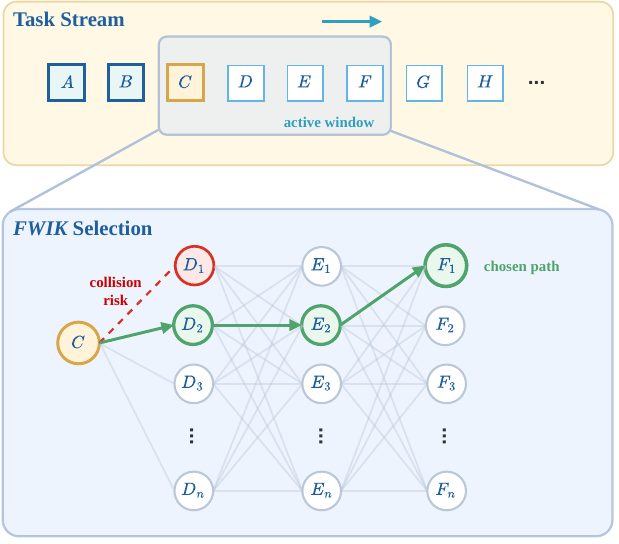}
    \caption{Overview of the \method decision pipeline.}
    \label{figure:OverallFWIK}
    \vspace{-5pt}
\end{figure}

\method is a short-lookahead decision layer for online multi-waypoint execution. At each control step, it selects an IK configuration for the next waypoint while accounting for its joint-space connectivity to nearby future waypoints. \method performs this decision in four stages: it defines a finite future window, generates IK candidates, ranks candidate paths through the window, and screens risky self-collision transitions before planner validation. Only the first goal of the selected path is executed; the window then advances and the decision is recomputed. Fig.~\ref{figure:OverallFWIK} summarizes the complete decision pipeline.

\subsection{Problem Formulation}

Let $q_{\mathrm{cur}}\in\mathbb{R}^{d}$ denote the current joint state of a manipulator with $d$ degrees of freedom. The ordered task contains $N$ Cartesian end-effector poses,
$\mathcal{T}
=
{T_0,T_1,\ldots,T_{N-1}}$,
where each $T_i$ denotes a Cartesian end-effector pose represented by position $[x,y,z]$ and unit-quaternion orientation $[q_x,q_y,q_z,q_w]$ in the planning frame.

At control step $t$, \method observes the finite future window
\begin{equation}
\mathcal{W}_t
=
{T_t,T_{t+1},\ldots,T_{\min(t+w-1,N-1)}},
\end{equation}
where $w$ is the maximum number of visible waypoints. The default setting is $w=3$. Rather than committing to a complete configuration sequence, \method selects one configuration for $T_t$, executes the planner-validated trajectory to that configuration, and then shifts the window forward by one waypoint. This receding-window execution limits commitment to the current action while allowing the IK decision to be revised as the task progresses.

\subsection{Candidate Generation and Window Graph}
For each pose $T_{t+r}\in\mathcal{W}_t$, the TRAC-IK solver is invoked through the MoveIt IK interface from randomized initial configurations, or seeds, to construct a candidate layer
\begin{equation}
Q_{t+r}
=
{q_{t+r}^{1},
q_{t+r}^{2},
\ldots,
q_{t+r}^{K_{t+r}}},
\qquad
q_{t+r}^{i}\in\mathbb{R}^{d}.
\end{equation}
Each retained candidate is an IK solution satisfying the target pose and active joint limits. Solutions outside the joint bounds are discarded. Candidate layers and adjacent edge-cost matrices are cached by waypoint identifier so that overlapping future windows can reuse previously computed information. In the reported experiments, \method requests up to $K=50$ candidates per waypoint, allows 800 IK attempts, and uses a $1$~ms timeout per attempt.

The resulting candidate layers form a directed layered graph. Each layer corresponds to one visible waypoint; each node corresponds to one IK candidate; and edges connect candidates in adjacent layers. An edge therefore represents a possible joint-space transition between two consecutive task poses.

For an edge from $q_{\mathrm{from}}$ to $q_{\mathrm{to}}$, the displacement of joint $m$ is $\Delta q_m
=
\left|
q_{\mathrm{to},m}
-
q_{\mathrm{from},m}
\right|$. \method assigns each edge a low-cost estimate of synchronized rest-to-rest motion time. Let $v_{\max,m}$ and $a_{\max,m}$ denote the velocity and acceleration limits of joint $m$. Under a symmetric acceleration--deceleration profile, the critical displacement required to reach the velocity limit is
\begin{equation}
d_{\mathrm{crit},m}
=
\frac{(v_{\max,m})^2}{a_{\max,m}}.
\end{equation}

If $\Delta q_m \leq d_{\mathrm{crit},m}$,
the joint follows a triangular profile: it accelerates and then decelerates without reaching its velocity limit. Its estimated motion time is
\begin{equation}
\tau_m
=
2
\sqrt{
\frac{\Delta q_m}{a_{\max,m}}
}.
\end{equation}

Otherwise, the joint reaches its velocity limit and follows a trapezoidal profile with an intermediate constant-velocity phase:
\begin{equation}
\tau_m
=
\frac{\Delta q_m}{v_{\max,m}}
+
\frac{v_{\max,m}}{a_{\max,m}}.
\end{equation}

Assuming synchronized joint motion with independent velocity and acceleration bounds, the transition duration is approximated by the slowest joint:
$\ell(q_{\mathrm{from}},q_{\mathrm{to}})
=
\max_m \tau_m$.

This cost does not generate a trajectory. Instead, it provides a computationally inexpensive and physically interpretable estimate for comparing IK branches before invoking a full motion planner~\cite{kroger2009online,kunz2012time}.

\subsection{Future-Window Path Ranking}

A path through the current future window selects one IK candidate from each visible waypoint layer:
\begin{equation}
\pi
=
\left(
q_t^{\pi_0},
q_{t+1}^{\pi_1},
\ldots,
q_{t+h-1}^{\pi_{h-1}}
\right),
\end{equation}
where
\begin{equation}
h = |\mathcal{W}_t| \leq w.
\end{equation}

The predicted traversal cost of a path is
\begin{equation}
J(\pi)
=
\ell\left(q_{\mathrm{cur}},q_t^{\pi_0}\right)
+
\sum_{r=0}^{h-2}
\ell\left(
q_{t+r}^{\pi_r},
q_{t+r+1}^{\pi_{r+1}}
\right).
\end{equation}
The first term estimates the transition from the current manipulator state to the next waypoint. The remaining terms estimate the cost of continuing through the nearby future waypoints. The objective therefore converts pointwise IK selection into a short-horizon path-ranking problem: two configurations that are similarly attractive for the immediate waypoint can receive different scores because they induce different transition costs to subsequent IK branches.

A direct enumeration of all candidate combinations would require up to $K^h$ paths when each layer contains $K$ candidates. Instead, the implementation propagates low-cost path suffixes backward through the layered graph and retains at most $R=8$ suffixes for each first-layer candidate. Consequently, at most $KR$ complete paths are retained for subsequent ranking. These paths are sorted in ascending order of $J(\pi)$.

Retaining multiple alternatives is important because the lowest-cost path according to the time model may later be rejected by the self-collision filter or by the downstream motion planner, while a slightly higher-cost alternative may remain feasible.

\subsection{Edge-Level Self-Collision Screening}

The time-based ranking model does not account for self-collision. A path with a low predicted traversal time may therefore begin with a transition that is unlikely to be executable. \method addresses this issue by screening the leading edges of each ranked path with a learned edge-level self-collision model before invoking the configured motion planner~\cite{zhi2022diffco,koptev2023neuraljsdf,li2024rdf}.

For each ranked path $\pi$, the current joint state is prepended to form
$P_\pi
=
\left(
q_{\mathrm{cur}},
q_t^{\pi_0},
q_{t+1}^{\pi_1},
\ldots,
q_{t+h-1}^{\pi_{h-1}}
\right)$.
\method evaluates the first $c$ adjacent transitions in $P_\pi$, where $c=2$ in the main experiments. Each transition is represented by its two endpoint configurations,
$e=[q_a,q_b]
\in
\mathbb{R}^{2\times d}$
and the model predicts
$p_{\mathrm{col}}
=
\sigma\!\left(f_\theta(e)\right),$
where $p_{\mathrm{col}}$ denotes the predicted probability of self-collision along the transition.

After per-joint normalization, the classifier uses a two-layer, four-head Transformer encoder with a 128-dimensional hidden representation~\cite{vaswani2017attention}. The Panda model, for example, was trained on one million independently sampled endpoint pairs. For each pair, Pilz PTP was used to generate a transition trajectory, and the returned trajectory was checked for self-collision using a maximum interpolation step of $0.02$~rad. These checks provided the binary training labels.

While training labels are extracted from Pilz PTP reference trajectories, the learned model is deployed as a pre-planning collision filter for all motion planners. It favors IK goals whose transitions are predicted to avoid self-collision, aiming to reduce detours and planned motion time. This preference may reject goals that another planner could reach through a detour; final trajectory validation remains with the configured planner. By assessing edge-level self-collisions, we explicitly filter for IK solutions enabling direct execution, balancing collision safety with trajectory efficiency.

The dataset was divided into 70\% training, 20\% validation, and 10\% testing subsets. At the validation-selected decision threshold of $0.682$, the held-out test set achieved an $F_1$ score of $0.972$ and an AUROC of $0.9988$. To be more conservative against self-collisions, online \method screening uses the threshold $\tau = 0.68$ and runs inference on CPU. A candidate path is rejected if any screened edge satisfies $p_{\mathrm{col}} \geq \tau$.

Edge predictions are cached within each control step. In particular, if the first edge associated with a first-layer IK candidate is predicted to collide, every retained path sharing that same first goal can be pruned without repeated inference.

When all candidate paths fail the default two-edge screening, \method repeats the screening using only the first edge. This fallback follows the receding-window execution policy: only the first transition is executed during the current control step. It therefore avoids rejecting a potentially safe immediate action solely because the predicted risk of the unexecuted second transition is uncertain.

The surviving paths are visited in ascending order of $J(\pi)$, while duplicate first-layer goals are skipped. Each distinct first IK target is submitted to the configured MoveIt 2 planner. \method returns the first target for which the planner produces a valid trajectory. The learned collision model therefore acts only as an inexpensive pre-planning risk filter rather than a safety proof; final geometric trajectory validation remains the responsibility of the downstream planner.

Fig.~\ref{alg:fwik} summarizes the complete decision procedure for one
\method control step, from candidate reuse and path ranking to edge screening
and planner validation.

\begin{figure}[t]
\hrule
\vspace{0.4em}
\begin{algorithmic}[1]
\REQUIRE
$q_{\mathrm{cur}}$,
future window $\mathcal{W}_t$,
IK generator,
optional edge-collision model,
planner
    \STATE Generate or reuse IK candidate layers $Q_{t+r}$.
    \STATE Build or reuse adjacent edge-time cost matrices.
    \STATE Retain up to $R$ low-cost suffixes per first-layer candidate.
    \STATE Rank the resulting candidate paths by $J(\pi)$.

    \IF{edge screening is enabled}
        \FOR{each unpruned ranked path $\pi$}
            \STATE Form $P_\pi$ by prepending $q_{\mathrm{cur}}$.
            \STATE Check the first $c$ edges of $P_\pi$.
            \STATE Reject $\pi$ if any checked edge is predicted colliding.
            \STATE Cache edge predictions.
            \STATE Prune paths sharing a colliding first edge.
        \ENDFOR

        \IF{all paths are rejected and $c>1$}
            \STATE Repeat screening using only the first edge.
        \ENDIF
    \ENDIF

    \STATE Visit surviving paths in ranked order.
    \STATE Skip duplicate first-layer IK goals.
    \STATE Return the first goal with a valid planner trajectory.
\end{algorithmic}
\vspace{0.4em}
\hrule
\caption{Decision process for one \method control step.}
\label{alg:fwik}
\vspace{-5pt}
\end{figure}

\section{Evaluation}
The evaluation measures complete waypoint sequences rather than isolated IK queries. This distinction is necessary because one poor intermediate configuration can stop an otherwise successful sequence~\cite{yoon2023learning,monari2023locally,xie2025many}. We first define the benchmark and metrics, then report the main comparison and an ablation study.

\subsection{Experimental Setup}

The software benchmark uses obstacle-free scenes and two 7-DoF manipulators: a Franka Emika Panda and a KUKA LBR iiwa. All tests run under Robot Operating System 2 (ROS 2) Jazzy and MoveIt 2. Each end-effector pose is obtained through forward kinematics, which computes the end-effector pose from a joint configuration. The sampled configuration is collision-free and away from kinematic singularities, where small end-effector motions can require large joint motions\cite{siciliano2009robotics}. The resulting sequences contain diverse goals and are not intended to represent smooth Cartesian paths. Every method uses the same 100 sequences, each with 50 waypoints. Each method-planner setting is repeated five times. A sequence counts as successful only if all 50 waypoints are completed. The final pose must be within $5~mm$ and $2^{\circ}$ of the target.

We compare \method with IKLink, STAMPEDE using adaptive sample expansion, and TRAC-IK in retry mode. The same downstream planner follows every IK method. The four planners are Pilz PTP, RRT-Connect, CHOMP, and STOMP. \method uses $w=3$, $c=2$, $K=50$, and $R=8$. TRAC-IK performs up to 50 retries until the motion planner outputs a feasible path. Both IKLink and STAMPEDE are offline full-path solvers that process the entire sequence as input and output IK solutions for all waypoints in a single pass. 

All experiments were conducted on a Mac equipped with an Apple M4 Pro chip and 24 GB of unified memory. Decision-time values support comparisons within this testbed, but they should not be treated as portable latency measurements.

\subsection{Metrics}
Sequence success is the main reliability metric. Waypoint completion is the fraction of waypoints completed before a sequence stops. Decision time includes IK generation, path ranking, collision screening, and planner validation. Planned trajectory duration is the sum of the durations of all planning steps in a completed sequence. MoveIt’s Time-Optimal Trajectory Generation (TOTG)~\cite{kunz2012time} response adapter determines each step duration by time-parameterizing the returned path under the configured joint velocity and acceleration limits. The resulting duration is planner-derived and does not represent measured robot execution or wall-clock time. Timing metrics include only completed sequences and must therefore be interpreted together with success rates. Reliability values report the mean and sample standard deviation (SD) across five repetitions. Timing values use the completed sequences that contain valid records. The study reports descriptive comparisons and does not claim statistical significance.

\begin{table*}[t]
\centering
\caption{Results over five repetitions. Reliability is mean$\pm$SD. Timing is mean$\pm$SD over completed sequences.}
\vspace{-3pt}
\label{tab:main_results}
\footnotesize
\setlength{\tabcolsep}{3.2pt}
\renewcommand{\arraystretch}{0.74}

\renewcommand\theadfont{\footnotesize}
\renewcommand\theadgape{\Gape[0pt]}

\newcommand{\pmA}[2]{\makebox[2.2em][r]{#1}\,$\pm$\,\makebox[1.4em][l]{#2}}
\newcommand{\pmB}[2]{\makebox[2.7em][r]{#1}\,$\pm$\,\makebox[1.8em][l]{#2}}
\newcommand{\pmC}[2]{\makebox[2.4em][r]{#1}\,$\pm$\,\makebox[2.2em][l]{#2}}
\newcommand{\pmD}[3][]{#1{\makebox[2.6em][r]{#2}\,$\pm$\,\makebox[1.8em][l]{#3}}}

\begin{tabular}{@{}cllcccc@{}}
\toprule
Robot & Planner & Method & \thead{Sequence Success\\(\%)} & \thead{Waypoint Completion\\(\%)} & \thead{Decision Time\\(s/sequence)} & \thead{Planned Duration\\(s/sequence)} \\
\midrule
\multirow{16}{*}{\rotatebox{90}{\textbf{Franka Emika Panda}}}
& \multirow{4}{*}{Pilz PTP}
& IKLink    & \pmA{10.4}{3.4}  & \pmB{40.06}{3.92} & \pmC{3.84}{0.24}  & \pmD{95.24}{4.29} \\
& & STAMPEDE  & \pmA{21.6}{3.6}  & \pmB{57.48}{2.13} & \pmC{3.89}{2.77}  & \pmD{84.49}{2.67} \\
& & TRAC-IK   & \pmA{9.6}{2.1}   & \pmB{41.79}{2.61} & \pmC{0.11}{0.02}  & \pmD{101.40}{4.80} \\
& & \method   & \pmA{96.8}{2.0}  & \pmB{98.19}{1.26} & \pmC{1.60}{0.12}  & \pmD[\textbf]{73.52}{1.61} \\
\cmidrule(lr){2-7}
& \multirow{4}{*}{RRT-Connect}
& IKLink    & \pmA{14.2}{4.4}  & \pmB{43.71}{3.19} & \pmC{3.77}{0.28}  & \pmD{96.24}{4.16} \\
& & STAMPEDE  & \pmA{19.8}{2.3}  & \pmB{54.88}{1.04} & \pmC{3.51}{1.90}  & \pmD{84.60}{2.41} \\
& & TRAC-IK   & \pmA{100.0}{0.0} & \pmB{100.00}{0.00}& \pmC{0.38}{0.07}  & \pmD{102.36}{4.32} \\
& & \method   & \pmA{100.0}{0.0} & \pmB{100.00}{0.00}& \pmC{1.83}{0.21}  & \pmD[\textbf]{73.54}{1.65} \\
\cmidrule(lr){2-7}
& \multirow{4}{*}{CHOMP}
& IKLink    & \pmA{11.6}{2.7}  & \pmB{41.12}{2.62} & \pmC{3.82}{0.26}  & \pmD{94.87}{3.88} \\
& & STAMPEDE  & \pmA{18.4}{2.6}  & \pmB{51.62}{1.91} & \pmC{3.34}{1.38}  & \pmD{84.41}{2.36} \\
& & TRAC-IK   & \pmA{93.2}{2.6}   & \pmB{99.82}{0.05} & \pmC{10.58}{2.49} & \pmD{104.78}{3.90} \\
& & \method   & \pmA{100.0}{0.0} & \pmB{100.00}{0.00}& \pmC{10.03}{1.19} & \pmD[\textbf]{73.48}{1.63} \\
\cmidrule(lr){2-7}
& \multirow{4}{*}{STOMP}
& IKLink    & \pmA{13.2}{1.9}  & \pmB{43.34}{1.04} & \pmC{3.82}{0.34}  & \pmD{95.13}{4.53} \\
& & STAMPEDE  & \pmA{19.4}{3.2}  & \pmB{54.63}{1.77} & \pmC{3.46}{2.05}  & \pmD{84.60}{2.32} \\
& & TRAC-IK   & \pmA{89.8}{1.9}   & \pmB{95.22}{0.97} & \pmC{26.67}{20.80}& \pmD{105.24}{4.29} \\
& & \method   & \pmA{100.0}{0.0} & \pmB{100.00}{0.00}& \pmC{2.11}{0.24}  & \pmD[\textbf]{73.53}{1.68} \\
\midrule
\multirow{16}{*}{\rotatebox{90}{\textbf{KUKA LBR iiwa}}}
& \multirow{4}{*}{Pilz PTP}
& IKLink    & \pmA{87.6}{2.3}  & \pmB{93.82}{1.04} & \pmC{2.25}{0.19}  & \pmD{104.02}{5.18} \\
& & STAMPEDE  & \pmA{85.6}{3.6}  & \pmB{92.20}{2.29} & \pmC{1.96}{0.48}  & \pmD{89.05}{2.87} \\
& & TRAC-IK   & \pmA{100.0}{0.0} & \pmB{100.00}{0.00}& \pmC{0.13}{0.01}  & \pmD{123.17}{5.29} \\
& & \method   & \pmA{100.0}{0.0} & \pmB{100.00}{0.00}& \pmC{1.27}{0.06}  & \pmD[\textbf]{77.61}{1.61} \\
\cmidrule(lr){2-7}
& \multirow{4}{*}{RRT-Connect}
& IKLink    & \pmA{100.0}{0.0} & \pmB{100.00}{0.00}& \pmC{2.24}{0.11}  & \pmD{103.97}{5.38} \\
& & STAMPEDE  & \pmA{100.0}{0.0} & \pmB{100.00}{0.00}& \pmC{1.91}{0.41}  & \pmD{89.14}{2.89} \\
& & TRAC-IK   & \pmA{100.0}{0.0} & \pmB{100.00}{0.00}& \pmC{0.33}{0.15}  & \pmD{121.07}{5.54} \\
& & \method   & \pmA{100.0}{0.0} & \pmB{100.00}{0.00}& \pmC{1.53}{0.22}  & \pmD[\textbf]{77.69}{1.62} \\
\cmidrule(lr){2-7}
& \multirow{4}{*}{CHOMP}
& IKLink    & \pmA{100.0}{0.0} & \pmB{100.00}{0.00}& \pmC{2.25}{0.12}  & \pmD{104.41}{5.32} \\
& & STAMPEDE  & \pmA{99.8}{0.4}  & \pmB{100.00}{0.01}& \pmC{2.04}{0.66}  & \pmD{89.11}{2.94} \\
& & TRAC-IK   & \pmA{100.0}{0.0} & \pmB{100.00}{0.00}& \pmC{10.01}{1.49} & \pmD{123.70}{5.54} \\
& & \method   & \pmA{100.0}{0.0} & \pmB{100.00}{0.00}& \pmC{10.81}{1.16} & \pmD[\textbf]{77.70}{1.60} \\
\cmidrule(lr){2-7}
& \multirow{4}{*}{STOMP}
& IKLink    & \pmA{97.6}{2.3}  & \pmB{98.38}{1.48} & \pmC{2.26}{0.13}  & \pmD{104.02}{5.20} \\
& & STAMPEDE  & \pmA{98.4}{0.9}  & \pmB{98.98}{0.51} & \pmC{1.92}{0.43}  & \pmD{89.03}{2.85} \\
& & TRAC-IK   & \pmA{100.0}{0.0} & \pmB{100.00}{0.00}& \pmC{0.49}{0.03}  & \pmD{123.43}{5.69} \\
& & \method   & \pmA{100.0}{0.0} & \pmB{100.00}{0.00}& \pmC{1.57}{0.11}  & \pmD[\textbf]{77.66}{1.55} \\
\bottomrule
\end{tabular}
\vspace{-3pt}
\end{table*}

\subsection{Main Results}

Table~\ref{tab:main_results} gives the complete comparison. Three findings are consistent across the benchmark. \method attains the highest sequence success in every robot-planner setting (including ties), yields the shortest planned trajectory in all eight settings, and generally spends more time in the selection stage than retry-based TRAC-IK.

The clearest reliability gain appears on the Panda, where baseline selectors often terminate early. \method completes 96.8--100\% of Panda sequences, whereas IKLink and STAMPEDE remain below one quarter in the same settings and retry-based TRAC-IK is sensitive to the planner. On the KUKA, most selectors are already close to the reliability ceiling, so the principal contribution there is efficiency rather than additional sequence completion. The difference between sequence success and waypoint completion in the harder Panda settings also confirms that a small number of late failures can dominate a 50-waypoint task.

Planned trajectory duration shows a consistent difference across both robots. \method shortens planned trajectories by 12.8--13.1\% relative to STAMPEDE, 22.5--25.6\% relative to IKLink, and 27.5--37.2\% relative to TRAC-IK (see representative execution in Fig.~\ref{fig:fwik_tracik_motion_sequence}). The small change across planners for the same robot is consistent with the purpose of the future-window cost, which selects joint configurations before the planner constructs a trajectory. The data support lower planned duration in this benchmark, but they do not establish global time optimality.

Decision time exposes the main cost of \method. TRAC-IK is usually faster because it returns a single solution, whereas \method generates alternatives, evaluates edges, and may validate multiple distinct goals. Both methods incur relatively large decision times under CHOMP, whose iteration typically entails computing functional gradients and updating the entire trajectory~\cite{zucker2013chomp}. For \method and retry-based TRAC-IK, the reported times include planner validation and retries, thus reflecting the combined computational characteristics of the planner and IK selection. Furthermore, since IKLink and STAMPEDE are offline, full-path methods, the decision times reported in Table~\ref{tab:main_results} correspond to the entire sequence. Notably, \method achieves average decision time per completed waypoint of $29.39 \pm 4.54\,\text{ms}$ (Pilz), $33.99 \pm 5.33\,\text{ms}$ (RRT-Connect), and $37.32 \pm 6.79\,\text{ms}$ (STOMP), demonstrating its suitability for online planning.

Overall, \method generally trades additional online computation for shorter planned duration and higher sequence completion. Because timing values include only completed sequences, they cannot be compared independently of reliability. This trade-off is favorable when motion efficiency matters more than a strict per-waypoint latency budget, but it may be unsuitable for high-rate loops. The current evidence therefore supports consistent motion-duration reduction and stable execution across the tested planners, rather than universal superiority across all metrics.

\begin{table}[t]
\centering
\caption{Ablation results on future window $w$ and edge screening.}
\label{tab:ablation}
\vspace{-3pt}
\scriptsize
\setlength{\tabcolsep}{2.5pt} 
\renewcommand{\arraystretch}{0.74}

\newcommand{\ablationcell}[2]{\makebox[3.3em][r]{#1}\,/\,\makebox[2.8em][l]{#2}}

\begin{tabular}{@{}lcccc@{}}
\toprule
\multirow{2}{*}{\raisebox{-0.7ex}{Planner}} & \multicolumn{2}{c}{$w=3$} & \multicolumn{2}{c}{$w=1$} \\
\cmidrule(lr){2-3} \cmidrule(lr){4-5}
& EC & no EC & EC$^*$ & no EC \\
\midrule
\multicolumn{5}{@{}l}{\textbf{Franka Panda}} \\
\addlinespace[1pt]
Pilz PTP        & \ablationcell{96.8\%}{73.52}  & \ablationcell{22.6\%}{73.22} & \ablationcell{92.0\%}{77.34} & \ablationcell{18.2\%}{77.22} \\
RRT-Connect     & \ablationcell{100.0\%}{73.54} & \ablationcell{100.0\%}{73.68}& \ablationcell{93.6\%}{77.43} & \ablationcell{100.0\%}{77.79} \\
CHOMP           & \ablationcell{100.0\%}{73.48} & \ablationcell{96.2\%}{73.37} & \ablationcell{95.8\%}{77.35} & \ablationcell{95.6\%}{77.43} \\
STOMP           & \ablationcell{100.0\%}{73.53} & \ablationcell{80.6\%}{73.43} & \ablationcell{93.4\%}{77.40} & \ablationcell{82.0\%}{77.45} \\
\midrule
\multicolumn{5}{@{}l}{\textbf{KUKA LBR iiwa}} \\
\addlinespace[1pt]
Pilz PTP        & \ablationcell{100.0\%}{77.61} & \ablationcell{100.0\%}{77.64}& \ablationcell{100.0\%}{81.80}& \ablationcell{100.0\%}{81.80} \\
RRT-Connect     & \ablationcell{100.0\%}{77.69} & \ablationcell{100.0\%}{77.62}& \ablationcell{100.0\%}{81.83}& \ablationcell{100.0\%}{81.86} \\
CHOMP           & \ablationcell{100.0\%}{77.70} & \ablationcell{100.0\%}{77.62}& \ablationcell{100.0\%}{81.85}& \ablationcell{100.0\%}{81.95} \\
STOMP           & \ablationcell{100.0\%}{77.66} & \ablationcell{100.0\%}{77.72}& \ablationcell{100.0\%}{81.89}& \ablationcell{100.0\%}{81.88} \\
\bottomrule
\end{tabular}

\vspace{2pt}
\parbox{\columnwidth}{\scriptsize EC denotes edge-collision screening. Each entry reports mean sequence success~(\%) / trajectory duration over completed sequences~(s) over 5 runs. RRT-Connect is evaluated via OMPL. $^*$The collision detection window $c$ is set to $1$ when $w=1$.}
\vspace{-10pt}
\end{table}

The ablation study in Table~\ref{tab:ablation} separates the effects of the future window and edge screening. With screening enabled, expanding the window from $w=1$ to $w=3$ reduces the planned duration by about 5\% across all robot-planner combinations. This supports the intended role of future-window reasoning: the next IK goal is chosen not only for its immediate transition cost, but also for the short sequence of transitions that it enables afterward. 

Screening primarily improves reliability when failures are common: on Panda-Pilz, it raises success from 22.6\% to 96.8\%, whereas its effect on the near-ceiling KUKA settings is negligible. This high baseline reliability can be largely attributed to the robot's highly symmetric 7-DoF kinematic layout and streamlined outer geometry with internally routed cables~\cite{bischoff2010kuka}, which structurally minimize interference contours and physical self-collisions compared to asymmetric platforms like the Panda. The Panda-RRT-Connect result at $w=1$, where screening lowers success from 100\% to 93.6\%, shows that false positives or an overly conservative threshold can remove viable branches. 

Edge screening functions as a risk-mitigation mechanism at the sequence level. Although it sacrifices a small fraction of potentially viable candidates, it significantly prevents self-colliding states from propagating to subsequent waypoints. Taken together, the ablation suggests that future-window reasoning primarily improves motion efficiency, whereas edge-level screening primarily improves reliability when collision-related failures are common.

\section{Discussion}

In this section, we analyze the performance mechanism of \method and discuss its current limitations and future potential.

\subsection{Empirical Insights and Trade-offs}
The evidence supports a specific claim: bounded lookahead makes online IK selection more effective at the sequence level. The most stable outcome is the reduction in trajectory duration across all eight settings. Reliability gains are conditional on the robot-planner pair, with the largest improvement on the Panda. On the KUKA, \method mainly preserves reliability while improving motion efficiency.

The ablation separates the mechanisms behind this behavior. A larger window improves efficiency by preferring IK branches that remain cheap over several upcoming transitions. Edge screening addresses a different failure mode by removing leading transitions that are likely to self-collide before planner retries are spent. Its benefit is therefore concentrated in failure-prone settings, while the observed false-positive case shows that screening thresholds can also discard a viable branch. The two components should consequently be tuned for different objectives rather than treated as interchangeable sources of improvement.

The main trade-off is computation versus downstream motion quality. \method spends more time before execution to generate alternatives, screen transitions, and validate goals. The extra cost is most defensible when planned motion duration and sequence completion dominate the application objective. It may be unsuitable for high-rate loops with a hard per-step budget. Because the study reports planner-derived duration rather than physical execution time, the observed reduction should not be interpreted as a direct wall-clock task-time improvement.

\subsection{Limitations and Future Work}
First, the evaluation is software-only, obstacle-free, and limited to two 7-DoF robots. Physical execution, environmental collision, controller tracking, and sensing uncertainty remain untested. Second, the benchmark waypoints are independently sampled collision-free, nonsingular poses rather than smooth trajectories from a particular application. This design intentionally stresses IK-branch changes and difficult transitions, but it does not represent all welding, painting, inspection, or pick-and-place tasks. Future evaluation should therefore include structured application trajectories as well as obstacle-rich scenes. Third, the learned edge self-collision classifier is robot-specific, requiring a separate model to be trained for each manipulator type. Its transferability across robots with different kinematic structures has not yet been evaluated. The degradation observed for Panda-RRT-Connect at $w=1$ indicates that screening can reduce reliability; its sensitivity to threshold selection warrants further study.

The current formulation also assumes that future waypoints are known exactly. This assumption is reasonable for predefined task paths but becomes restrictive when future targets are obtained from perception or prediction. In that setting, later graph layers could incorporate pose uncertainty or be weighted by prediction confidence. A further extension would combine this uncertainty-aware window with scene-conditioned distance models and self-collision estimates~\cite{koptev2023neuraljsdf,murali2023cabinet}. Such extensions would preserve the receding-window structure of \method while allowing the decision layer to reason jointly about future connectivity, collision risk, and uncertainty.

\section{CONCLUSION}

In this work, we present \method, an online IK selector for ordered waypoint tasks. The method ranks IK paths over a small future window, filters risky transitions with a learned self-collision model, and leaves final trajectory validation to a standard motion planner. In software tests on two 7-DoF robots and four MoveIt~2 planners, \method completed 96.8--100\% of sequences and produced the shortest planned trajectory duration in all eight settings. The reliability improvement was largest on the Panda, while several KUKA settings were already close to 100\%. \method usually requires more decision time than retry-based TRAC-IK. Its main contribution is therefore a practical exchange: more work during IK selection for higher sequence completion and shorter planned motion. The results support future-window selection in this benchmark and define the physical, statistical, and scene-level tests needed for broader conclusions.

\bibliographystyle{IEEEtran}

\balance
\bibliography{references}

@book{siciliano2009robotics,
title     = {{Robotics: Modelling, Planning and Control}},
author    = {Siciliano, Bruno and Sciavicco, Lorenzo and Villani, Luigi and Oriolo, Giuseppe},
publisher = {Springer},
year      = {2009},
doi       = {10.1007/978-1-84628-642-1}
}

@article{coleman2014moveit,
title   = {{Reducing the Barrier to Entry of Complex Robotic Software: A MoveIt! Case Study}},
author  = {Coleman, David and {\c{S}}ucan, Ioan A. and Chitta, Sachin and Correll, Nikolaus},
journal = {Journal of Software Engineering for Robotics},
volume  = {5},
number  = {1},
pages   = {3--16},
year    = {2014},
doi     = {10.6092/JOSER_2014_05_01_p3}
}

@article{sucan2012ompl,
title   = {{The Open Motion Planning Library}},
author  = {{\c{S}}ucan, Ioan A. and Moll, Mark and Kavraki, Lydia E.},
journal = {IEEE Robotics \& Automation Magazine},
volume  = {19},
number  = {4},
pages   = {72--82},
year    = {2012},
doi     = {10.1109/MRA.2012.2205651}
}

@inproceedings{beeson2015tracik,
title     = {{{TRAC-IK}: An Open-Source Library for Improved Solving of Generic Inverse Kinematics}},
author    = {Beeson, Patrick and Ames, Barrett},
booktitle = {Proceedings of the IEEE-RAS International Conference on Humanoid Robots},
pages     = {928--935},
year      = {2015},
doi       = {10.1109/HUMANOIDS.2015.7363472}
}

@article{nakamura1986singularity,
title   = {{Inverse Kinematic Solutions with Singularity Robustness for Robot Manipulator Control}},
author  = {Nakamura, Yoshihiko and Hanafusa, Hideo},
journal = {Journal of Dynamic Systems, Measurement, and Control},
volume  = {108},
number  = {3},
pages   = {163--171},
year    = {1986},
doi     = {10.1115/1.3143764}
}

@inproceedings{rakita2019stampede,
title     = {{{STAMPEDE}: A Discrete-Optimization Method for Solving Pathwise-Inverse Kinematics}},
author    = {Rakita, Daniel and Mutlu, Bilge and Gleicher, Michael},
booktitle = {Proceedings of the IEEE International Conference on Robotics and Automation},
pages     = {3507--3513},
year      = {2019},
doi       = {10.1109/ICRA.2019.8793617}
}

@inproceedings{wang2024iklink,
title     = {{{IKLink}: End-Effector Trajectory Tracking with Minimal Reconfigurations}},
author    = {Wang, Yeping and Sifferman, Carter and Gleicher, Michael},
booktitle = {Proceedings of the IEEE International Conference on Robotics and Automation},
pages     = {13629--13636},
year      = {2024},
doi       = {10.1109/ICRA57147.2024.10611510}
}

@inproceedings{rakita2018relaxedik,
  author    = {Daniel Rakita and Bilge Mutlu and Michael Gleicher},
  title     = {{{RelaxedIK}: Real-time Synthesis of Accurate and Feasible Robot Arm Motion}},
  booktitle = {Proceedings of Robotics: Science and Systems},
  year      = {2018},
  address   = {Pittsburgh, Pennsylvania},
  month     = jun,
  doi       = {10.15607/RSS.2018.XIV.043}
}

@article{kang2022torm,
    title   = {{Analysis and Acceleration of {TORM}: Optimization-Based Planning for Path-Wise Inverse Kinematics}},
    author  = {Kang, Mincheul and Yoon, Sung-Eui},
    journal = {Autonomous Robots},
    volume  = {46},
    number  = {5},
    pages   = {599--615},
    year    = {2022},
    doi     = {10.1007/s10514-022-10040-1}
}

@article{alatartsev2015task,
title   = {{Robotic Task Sequencing Problem: A Survey}},
author  = {Alatartsev, Sergey and Stellmacher, Sebastian and Ortmeier, Frank},
journal = {Journal of Intelligent \& Robotic Systems},
volume  = {80},
number  = {2},
pages   = {279--298},
year    = {2015},
doi     = {10.1007/s10846-015-0190-6}
}

@inproceedings{suarez2018robotsp,
  title={{Robotsp--a fast solution to the robotic task sequencing problem}},
  author={Su{\'a}rez-Ruiz, Francisco and Lembono, Teguh Santoso and Pham, Quang-Cuong},
  booktitle={2018 IEEE International Conference on Robotics and Automation (ICRA)},
  pages={1611--1616},
  year={2018},
  organization={IEEE}
}

@article{kavraki1996prm,
title   = {{Probabilistic Roadmaps for Path Planning in High-Dimensional Configuration Spaces}},
author  = {Kavraki, Lydia E. and {\v{S}}vestka, Petr and Latombe, Jean-Claude and Overmars, Mark H.},
journal = {IEEE Transactions on Robotics and Automation},
volume  = {12},
number  = {4},
pages   = {566--580},
year    = {1996},
doi     = {10.1109/70.508439}
}

@misc{pilz2018industrial,
  author       = {{Pilz GmbH \& Co. KG}},
  title        = {Industrial Trajectory Generation for {MoveIt!}},
  year         = {2018},
  howpublished = {ROSIN Focused Technical Project},
  url          = {https://www.rosin-project.eu/ftp/industrial-trajectory-generation},
  note         = {Software and documentation, accessed Sep. 1, 2026}
}

@inproceedings{kuffner2000rrt,
  title={{RRT-connect: An efficient approach to single-query path planning}},
  author={Kuffner, James J and LaValle, Steven M},
  booktitle={Proceedings 2000 ICRA. Millennium conference. IEEE international conference on robotics and automation. Symposia proceedings (Cat. No. 00CH37065)},
  volume={2},
  pages={995--1001},
  year={2000},
  organization={IEEE}
}

@article{zucker2013chomp,
title   = {{{CHOMP}: Covariant Hamiltonian Optimization for Motion Planning}},
author  = {Zucker, Matthew and Ratliff, Nathan and Dragan, Anca D. and Pivtoraiko, Mihail and Klingensmith, Matthew and Dellin, Christopher M. and Bagnell, J. Andrew and Srinivasa, Siddhartha S.},
journal = {The International Journal of Robotics Research},
volume  = {32},
number  = {9-10},
pages   = {1164--1193},
year    = {2013},
doi     = {10.1177/0278364913488805}
}

@inproceedings{kalakrishnan2011stomp,
title     = {{{STOMP}: Stochastic Trajectory Optimization for Motion Planning}},
author    = {Kalakrishnan, Mrinal and Chitta, Sachin and Theodorou, Evangelos and Pastor, Peter and Schaal, Stefan},
booktitle = {Proceedings of the IEEE International Conference on Robotics and Automation},
pages     = {4569--4574},
year      = {2011},
doi       = {10.1109/ICRA.2011.5980280}
}

@article{schulman2014trajopt,
title   = {{Motion Planning with Sequential Convex Optimization and Convex Collision Checking}},
author  = {Schulman, John and Duan, Yan and Ho, Jonathan and Lee, Alex and Awwal, Ibrahim and Bradlow, Henry and Pan, Jia and Patil, Sachin and Goldberg, Ken and Abbeel, Pieter},
journal = {The International Journal of Robotics Research},
volume  = {33},
number  = {9},
pages   = {1251--1270},
year    = {2014},
doi     = {10.1177/0278364914528132}
}

@inproceedings{pan2012fcl,
title     = {{{FCL}: A General Purpose Library for Collision and Proximity Queries}},
author    = {Pan, Jia and Chitta, Sachin and Manocha, Dinesh},
booktitle = {Proceedings of the IEEE International Conference on Robotics and Automation},
pages     = {3859--3866},
year      = {2012},
doi       = {10.1109/ICRA.2012.6225337}
}

@article{das2020learning,
  title={{Learning-based proxy collision detection for robot motion planning applications}},
  author={Das, Nikhil and Yip, Michael},
  journal={IEEE Transactions on Robotics},
  volume={36},
  number={4},
  pages={1096--1114},
  year={2020},
  publisher={IEEE}
}

@article{kroger2009online,
  title={{Online trajectory generation: Basic concepts for instantaneous reactions to unforeseen events}},
  author={Kr{\"o}ger, Torsten and Wahl, Friedrich M},
  journal={IEEE Transactions on Robotics},
  volume={26},
  number={1},
  pages={94--111},
  year={2009},
  publisher={IEEE}
}

@inproceedings{kunz2012time,
title     = {{Time-Optimal Trajectory Generation for Path Following with Bounded Acceleration and Velocity}},
author    = {Kunz, Tobias and Stilman, Mike},
booktitle = {Robotics: Science and Systems},
year      = {2012}
}

@inproceedings{vaswani2017attention,
title     = {{Attention Is All You Need}},
author    = {Vaswani, Ashish and Shazeer, Noam and Parmar, Niki and Uszkoreit, Jakob and Jones, Llion and Gomez, Aidan N. and Kaiser, Lukasz and Polosukhin, Illia},
booktitle = {Advances in Neural Information Processing Systems},
volume    = {30},
pages     = {5998--6008},
year      = {2017}
}

@article{ames2022ikflow,
  title={{Ikflow: Generating diverse inverse kinematics solutions}},
  author={Ames, Barrett and Morgan, Jeremy and Konidaris, George},
  journal={IEEE Robotics and Automation Letters},
  volume={7},
  number={3},
  pages={7177--7184},
  year={2022},
  publisher={IEEE}
}

@article{wang2023rangedik,
  title={{Rangedik: An optimization-based robot motion generation method for ranged-goal tasks}},
  author={Wang, Yeping and Praveena, Pragathi and Rakita, Daniel and Gleicher, Michael},
  journal={arXiv preprint arXiv:2302.13935},
  year={2023}
}

@inproceedings{yoon2023learning,
  author    = {Yoon, Minsung and Kang, Mincheul and Park, Daehyung and Yoon, Sung-Eui},
  title     = {{Learning-Based Initialization of Trajectory Optimization for Path-Following Problems of Redundant Manipulators}},
  booktitle = {Proceedings of the IEEE International Conference on Robotics and Automation},
  pages     = {9686--9692},
  year      = {2023},
  doi       = {10.1109/ICRA48891.2023.10161426}
}

@inproceedings{monari2023locally,
  author    = {Monari, Eugenio and Chen, Yi and Vertechy, Rocco},
  title     = {{On Locally Optimal Redundancy Resolution Using the Basis of the Null Space}},
  booktitle = {Proceedings of the IEEE International Conference on Robotics and Automation},
  pages     = {9665--9671},
  year      = {2023},
  doi       = {10.1109/ICRA48891.2023.10161181}
}

@inproceedings{xie2025many,
  author    = {Xie, Bin and Zhao, Jiaming and Wang, Qingfeng and Wu, Di},
  title     = {{Many-Objective Motion Generation Method for Redundant Manipulators by Solving Pathwise Inverse Kinematics}},
  booktitle = {Proceedings of the IEEE/RSJ International Conference on Intelligent Robots and Systems},
  pages     = {5206--5213},
  year      = {2025},
  doi       = {10.1109/IROS60139.2025.11247737}
}

@inproceedings{morgan2024cppflow,
  title={{Cppflow: Generative inverse kinematics for efficient and robust cartesian path planning}},
  author={Morgan, Jeremy and Millard, David and Sukhatme, Gaurav S},
  booktitle={2024 IEEE International Conference on Robotics and Automation (ICRA)},
  pages={12279--12785},
  year={2024},
  organization={IEEE}
}

@article{oelerich2025boundmpc,
  author  = {Oelerich, Thies and Beck, Florian and Hartl-Nesic, Christian and Kugi, Andreas},
  title   = {{{BoundMPC}: Cartesian Path Following With Error Bounds Based on Model Predictive Control in the Joint Space}},
  journal = {The International Journal of Robotics Research},
  volume  = {44},
  number  = {8},
  pages   = {1287--1316},
  year    = {2025},
  doi     = {10.1177/02783649241309354}
}

@article{zhi2022diffco,
  title={{Diffco: Autodifferentiable proxy collision detection with multiclass labels for safety-aware trajectory optimization}},
  author={Zhi, Yuheng and Das, Nikhil and Yip, Michael},
  journal={IEEE Transactions on Robotics},
  volume={38},
  number={5},
  pages={2668--2685},
  year={2022},
  publisher={IEEE}
}

@inproceedings{li2024rdf,
  author    = {Li, Yiming and Zhang, Yan and Razmjoo, Amirreza and Calinon, Sylvain},
  title     = {{Representing Robot Geometry as Distance Fields: Applications to Whole-Body Manipulation}},
  booktitle = {Proceedings of the IEEE International Conference on Robotics and Automation},
  pages     = {15351--15357},
  year      = {2024},
  doi       = {10.1109/ICRA57147.2024.10611674}
}

@inproceedings{rakita2021collisionik,
  author    = {Rakita, Daniel and Shi, Haochen and Mutlu, Bilge and Gleicher, Michael},
  title     = {{{CollisionIK}: A Per-Instant Pose Optimization Method for Generating Robot Motions With Environment Collision Avoidance}},
  booktitle = {Proceedings of the IEEE International Conference on Robotics and Automation},
  pages     = {9995--10001},
  year      = {2021},
  doi       = {10.1109/ICRA48506.2021.9561505}
}

@inproceedings{murali2023cabinet,
  author    = {Murali, Adithyavairavan and Mousavian, Arsalan and Eppner, Clemens and Fishman, Adam and Fox, Dieter},
  title     = {{{CabiNet}: Scaling Neural Collision Detection for Object Rearrangement With Procedural Scene Generation}},
  booktitle = {Proceedings of the IEEE International Conference on Robotics and Automation},
  pages     = {1866--1874},
  year      = {2023},
  doi       = {10.1109/ICRA48891.2023.10161528}
}

@article{koptev2023neuraljsdf,
  author  = {Koptev, Mikhail and Figueroa, Nadia and Billard, Aude},
  title   = {{{Neural-JSDF}: Neural Joint Space Implicit Signed Distance Functions for Reactive Robot Manipulator Control}},
  journal = {IEEE Robotics and Automation Letters},
  volume  = {8},
  number  = {2},
  pages   = {480--487},
  year    = {2023},
  doi     = {10.1109/LRA.2022.3227860}
}

@inproceedings{bischoff2010kuka,
  title={{The KUKA-DLR Lightweight Robot arm-a new reference platform for robotics research and manufacturing}},
  author={Bischoff, Rainer and Kurth, Johannes and Schreiber, G{\"u}nter and Koeppe, Ralf and Albu-Sch{\"a}ffer, Alin and Beyer, Alexander and Eiberger, Oliver and Haddadin, Sami and Stemmer, Andreas and Grunwald, Gerhard and others},
  booktitle={ISR 2010 (41st international symposium on robotics) and ROBOTIK 2010 (6th German conference on robotics)},
  pages={1--8},
  year={2010},
  organization={VDE}
}

\end{document}